# Instrument-Armed Bandits

Nathan Kallus

School of Operations Research and Information Engineering and Cornell Tech,

Cornell University

New York, New York 10011

`kallus@cornell.edu`

**Abstract**

We extend the classic multi-armed bandit (MAB) model to the setting of noncompliance, where the arm pull is a mere instrument and the treatment applied may differ from it, which gives rise to the instrument-armed bandit (IAB) problem. The IAB setting is relevant whenever the experimental units are human since free will, ethics, and the law may prohibit unrestricted or forced application of treatment. In particular, the setting is relevant in bandit models of dynamic clinical trials and other controlled trials on human interventions. Nonetheless, the setting has not been fully investigate in the bandit literature. We show that there are various and divergent notions of regret in this setting, all of which coincide only in the classic MAB setting. We characterize the behavior of these regrets and analyze standard MAB algorithms. We argue for a particular kind of regret that captures the causal effect of treatments but show that standard MAB algorithms cannot achieve sublinear control on this regret. Instead, we develop new algorithms for the IAB problem, prove new regret bounds for them, and compare them to standard MAB algorithms in numerical examples.

# 1  Introduction

Multi-armed bandits (MABs) are often used to model dynamic clinical trials [18]. In a clinical trial interpretation of an MAB, an experimenter applies one of $m$ treatments to each incoming patient, the reward of the applied treatment is recorded, and the target is to learn what the best treatment is so as to ensure that the average of all patient rewards over the trial is near-optimal. However, when patients are human, free will, ethics, and the rule of law often prohibit completely liberal application of treatment by the experimenter. In particular, patients may choose not to comply with the treatment assigned. By some estimates, non-compliance rates in some clinical contexts may be as high as 50% [20]. In real clinical trials, subject non-compliance is generally present and can often constitute a significant effect. The same may occur in non-clinical dynamic randomized controlled trials (RCTs) interpretations of MABs, such as trials investigating policy and training interventions. Whenever the experimental unit is human, non-compliance may be a significant presence.

This departure from the standard MAB model has not been fully investigated in the bandit literature but nonetheless can have significant implications on both performance metrics and what constitutes a good bandit algorithm. In particular, almost all bandit models assume full control of treatment assignment or do consider perturbations to control (as in an MDP) but do not consider regret quantities that correspond to the causal effect of treatment and to bandit algorithms that are aimed at finding the treatment(s) with the best causal effect.

The purpose of this paper is to explore the implications of non-compliance for MABs and develop new theory and methods appropriate for these settings. Toward this end, we define the instrument-armed bandit problem where arms do not necessarily correspond to the treatment administered and establish a number of new regret quantities that are of unique interest in this setting. We characterize these regret quantities in various different settings of non-compliance and analyze the corresponding regret of standard MAB algorithms like the upper confidence bound (UCB) algorithm. We find that standard MAB algorithms can fail to achieve sublinear regret on the compliers (which we define) and can fail to learn what is the best treatment. We therefore develop new bandit algorithms that address this issue and prove new regret bounds for these algorithms. We conclude with a numerical investigation.

## 2   The Instrument-Armed Bandit Problem

In the presence of non-compliance, the assignment to treatment – or arm pull – by the experimenter is a mere instrument: it does not guarantee the application of the treatment but it may influence it. In the language of econometrics [2], the variable indicating assignment to treatment is an *instrumental variable*. Thus, in the presence of non-compliance, the pull of an arm corresponds only to the *intent to treat* (ITT) rather than the administration of treatment.

We call the bandit problem that arises in the presence of non-compliance the instrument-armed bandit (IAB) problem because each arm pull represents only the choice of instrument. We consider the IAB problem with $m$ arms where each arm represents both an instrument and a treatment. Unlike the standard MAB problem, the IAB problem involves two forms of bandit feedback: the reward of the treatment applied and the identity of the treatment applied, which we assume is always observed.

In the presence of non-compliance, whether a subject complies with an assigned treatment, that is, whether the treatment assigned is also applied, generally depends to a great extent on the treatment assigned. Some subjects may be averse to new, intense, or invasive treatments and reject these when assigned. Others may be more eager to try these treatments and may reject insufficiently-intensive treatments. Depending on the application, certain modes of non-compliance may be ruled out: in some settings a subject can demand any one treatment and in other settings a subject can only reject a new or invasive treatment but cannot demand it unless assigned to it. Assuming, as we do, that we observe the treatment (or non-treatment, for that matter) in-fact applied to the unit, the compliance of the unit is an incomplete observation and comprises a form of *bandit feedback*: we only know the compliance or non-compliance resulting from the treatment assigned. For example, a subject eager to try intensive treatment may appear to comply when given such an assignment and may appear not to comply when given another assignment.

We formulate this by saying that an experimental unit is characterized by a vector of $m$

integers $\chi = (\chi(1), \ldots, \chi(m))$, each taking values in $[m] = \{1, \ldots, m\}$ and each denoting the treatment in-fact applied if the respective arm is pulled. Thus, when we pull arm $z \in [m]$, the realized-treatment feedback we observe is $\chi(z)$. As with unrealized rewards in standard MAB problems, $\chi(z')$ for other arms $z'$ are not observed. We call $\chi$ the *compliance type* of an experimental unit.

A unit is said to be a *complier* if it complies with any treatment assigned. That is, letting $\iota = (1, \ldots, m)$ so that $\iota(z) = z$ is the identity mapping, a unit is complier whenever $\chi = \iota$. A unit is a *non-complier* if $\chi \neq \iota$. One particular type of non-compliers are always-takers: letting $\kappa^{(t)} = (t, \ldots, t)$ for $t \in [m]$ so that $\kappa^{(t)}(z) = t$ for all $z \in [m]$, a unit is said to be a *t-always-taker* if $\chi = \kappa^{(t)}$.

As in the standard MAB problem, each unit is also characterized by $m$ reward values $Y(1), \ldots, Y(m)$ taking values on the real line. When the treatment applied is $x$, the reward observed is $Y(x)$. We will assume that high rewards are preferable. The variables $\chi(1), Y(1), \ldots, \chi(m), Y(m)$ denote a generic draw of an experimental unit from a population of units.

In the IAB problem, at each round $t = 1, 2, \ldots$, an incoming unit given by $(\chi_t(1), Y_t(1), \ldots, \chi_t(m), Y_t(m))$ is drawn at random from the population of units, independently from past draws and past actions. With nothing about the unit revealed, the experimenter chooses an intended arm $Z_t \in [m]$. The treatment actually applied is $X_t = \chi_t(Z_t)$, which is revealed, and the reward collected is $Y_t = Y_t(\chi_t(Z_t))$, which is also revealed.[1] We summarize the IAB problem as:

> **The instrument-armed bandit problem**
> For each round $t = 1, 2, \ldots$
> 1. $\chi_t(1), Y_t(1), \ldots, \chi_t(m), Y_t(m)$ are drawn independently from the past and are hidden;
> 2. the experimenter pulls arm $Z_t \in [m]$;
> 3. $X_t = \chi_t(Z_t)$ and $Y_t = Y_t(\chi_t(Z_t))$ are revealed.

The parameters of the problem revealed to the experimenter at the onset are the number of arms $m$ and, potentially but not necessarily, the number of rounds $T$. The hidden parameter characterizing an instance of the IAB problem is the distribution of units (step 1 above). The distribution of $\chi$ and the conditional marginal distributions of $Y(x)$ given $\chi$ for each $x$ separately are denoted by

$$p_\kappa = \mathbb{P}(\chi = \kappa), \quad \nu_{x|\kappa}(A) = \mathbb{P}(Y(x) \in A \mid \chi = \kappa).$$

---

[1]Note that in our notation, $\chi_t = (\chi_t(1), \ldots, \chi_t(m))$ is a vector of potential treatments whereas $Y_t$ is a scalar realized reward and should not be confused with the vector $(Y_t(1), \ldots, Y_t(m))$ of potential rewards.

Figure 1: The Instrumental Variable Model

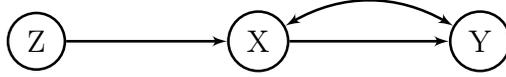

We assume that $\int |y|\, d\nu_{x|\kappa} < \infty$ for all $x \in [m], \kappa \in [m]^m$ and define:

$$\mu_{x|\kappa} = \mathbb{E}[Y(x) \mid \chi = \kappa] = \int y\, d\nu_{x|\kappa},\ \mu_x = \mathbb{E}[Y(x)] = \sum_{\kappa \in [m]^m} p_\kappa \mu_{x|\kappa},$$
$$\tilde{\mu}_z = \mathbb{E}[Y(\chi(z))] = \sum_{\kappa \in [m]^m} p_\kappa \mu_{\kappa(z)|\kappa}.$$

An instance of IAB is said to be an instance of MAB when $p_\iota = 1$, *i.e.*, when all units are compliers. In this case, the IAB problem reduces to the standard stochastic MAB problem with arm reward distributions $\nu_{x|\iota}$ [for stochastic MAB, see 7, p. 4].

A (randomized) bandit algorithm $\pi$ is a mapping from the step $t$ and the observed data up to time $t$, $\mathcal{D}_t = \{Z_1, X_1, Y_1, \ldots, Z_{t-1}, X_{t-1}, Y_{t-1}\}$, to a probability distribution over the arms $[m]$, which determines how we sample the arm pulled $Z_t$ in step 2 of the IAB problem. Equivalently, $\pi$ is an assignment to the random variable $Z_t$ in the random process defining the problem such that it is independent of present and future given the past. Under an algorithm $\pi$, the random process $(Z_1, X_1, Y_1), (Z_2, X_2, Y_2), \ldots$ is well defined and we can compute various cumulative expectations over it, which we will call regret. We emphasize that a probability or expectation is taken under the algorithm $\pi$ by $\mathbb{P}_\pi$ and $\mathbb{E}_\pi$, respectively.

In the IAB problem, whose setting can also be represented as a directed acyclic graph [16] (Fig. 1), there are various divergent notions of regret, which we discuss in Sec. 3. These divergent notions all coincide when the IAB problem is an instance of MAB, but are otherwise generally distinct.

## 2.1 Background and Related Literature

The IAB problem is an example of a *stochastic* bandit problem: the distributions of reward and, in the IAB case, compliance type are fixed and stationary but *unknown.* We focus solely on this stochastic setting. We refer the reader to Bubeck and Cesa-Bianchi [7] for a comparison of stochastic and adversarial settings in the classic MAB problem. Regret in the classic MAB problem is defined as the difference between the cumulative reward of always pulling the a priori best single arm (as per the unknown reward distributions) and the cumulative reward of the bandit policy up to a horizon $T$. When all units are compliers, the IAB problem reduces to the stochastic MAB, for which Lai and Robbins [11] in their seminal paper lower bounded the expected regret of any policy by the logarithm of $T$ times factors that depend on the difference between suboptimal and optimal arms. A minimal requirement of a bandit algorithm for MAB is that it be *order optimal* in that it achieve regret that has order no more than logarithmic in $T$. The popular UCB algorithm [4] is an example of one such order optimal algorithm. Recently, Agrawal and Goyal [1] proved that the posterior sampling algorithm of [17] is also order optimal. Other example include Garivier and Cappé [8], Kaufmann et al. [10], Maillard et al. [15].

A variant of the MAB problem is the MAB with latent class (MABLC) problem [14], in which the reward vector is *jointly* drawn with class membership and regret is defined as the difference between the cumulative reward of always pulling the a priori best arm given class and the cumulative reward of a bandit policy that is given an observation of a *finer* quantity than class (such as a unit identity) before pulling an arm. In the IAB problem, we can conceive of compliance type as a latent class, especially in light of the possibility of heterogeneous rewards, *i.e.*, we may have $\mu_{x|\kappa} \neq \mu_{x|\kappa'}$ for $\kappa \neq \kappa'$. Compliance type as a class does not influence any actual bandit algorithm for IAB because nothing indicative of compliance type is revealed *before* an arm pull. Nonetheless, we get different forms of IAB regret depending on whether we consider benchmark policies that may or may not be allowed to leverage a priori class knowledge. For the IAB problem, as we show in Sec. 3, in first case, there is no hope of sublinear regret and, in the second, regret may be negative, and so in fact an intermediate form of regret is needed, which considers latent class but restricts to a certain subpopulation of units.

Another variant of the MAB problem is the MAB with unobserved confounder (MABUC) problem [5], where a latent factor influences reward distributions and the experimenter, rather than being given a finer observation as in MABLC, is given a *coarser* observation of an "intuitive" arm choice, corresponding to a particular stationary joint distribution of latent factors, rewards, and "intuitive" arms. The authors show that when regret is measured relative to the best arm given the latent factor rather than marginal best choice, standard bandit algorithms that use only observations of pulled arm and resultant reward may achieve linear regret. Similarly, we will show that in the IAB problem, if one is not interested in the simple intent-to-treat regret (which we will define), then standard bandit algorithms achieve linear regret under most alternative notions of regret that consider causal effects. Lattimore et al. [12] consider a closely related problem to the MABUC without observation of an "intuitive" arm and focusing on the simple regret of eventually recommending the best arm.

Instrumental variable analysis is a method to estimate causal relationships between a possible treatment variable and an outcome variable in the absence of a fully randomized experiment that intervenes on the treatment in a fully controlled manner [2]. It relies on the availability of an *instrumental variable*, which may only affect the outcome via the treatment and is associated with the treatment. When all of these relationships are additive (*i.e.*, linear) the two-stage least squares (2SLS) procedure, which first regresses treatment on the instrument and then regresses the outcome on the result, provides a consistent estimate of causal effect. When this effect is heterogeneous in the population (*i.e.*, not simply additive), this estimate is in fact of a particular average effect known as the local-average treatment effect, or LATE [9].

In a clinical trial with non-compliance, a randomized ITT is an example of an instrumental variable. When treatment is binary (treatment vs control), 2SLS is equivalent to the Wald estimator [19], which is the ratio of a naïve estimate of the ITT effect, given by the difference in mean outcomes in units intended for treatment and for control, divided by an estimate of the number of compliers $p_\iota$, given by the difference in mean number receiving treatment among the two intended groups. When the population contains only compliers, treatment-always takers, and control-always takers, the Wald estimator is consistent for the LATE, which in this case is the average effect on the compliers [3, 13].

In the IAB problem, $Z_t$ corresponds to the ITT for unit $t$ and, as such, is an example of an instrumental variable. The treatment may be more than binary as there may be more than two arms ($m \geq 2$) and $Z_t$ is chosen dynamically for each $t$ by the bandit algorithm as rounds progress. The question we ask is how to best choose $Z_t$ so to balance exploration and exploitation. The right way to measure how well we do this hinges on what is our goal. If it is to get at the best causal effect of a treatment then we must consider new forms of regret that do not exist in the vanilla MAB context. In the next section, we develop several new notions of regret for the IAB problem and characterize their behavior.

## 3 IAB Regret

The target of a bandit algorithm is to achieve low regret. Unlike the MAB problem, in the case of IAB there are a wide range of candidate definitions for regret, with different meanings. In the case that an instance of IAB is an instance of MAB ($p_\iota = 1$), however, we show that all of these notions of regret in fact coincide with the classic notion of MAB regret. In this section, we present the various notions of regret and characterize their behavior.

### 3.1 Intent-to-Treat Regret

In the case that one is not interested in the causal effect of treatment and only in the observational resultant rewards of the ITT instrument, then the IAB problem can easily be reduced to an MAB problem. We simply ignore the instrumental aspect of the arms and ignore all bandit feedback of non-compliance. This leaves us with only arm pulls and resultant rewards, disregarding the mechanism by which the reward was begotten. If we do this, then the only relevant notion of regret is the MAB regret, which measures how much regret we have in pulling one instrument arm compared to another instrument arm, rather than compared to any other possible treatment.

In the IAB problem, we call this the ITT regret. To define ITT regret, let $z^*$ be an arbitrary choice in the set of *marginally* optimal instrument arms:

$$z^* \in \mathcal{Z}^* = \mathrm{argmax}_{z \in [m]} \tilde{\mu}_z, \quad \tilde{\mu}^* = \tilde{\mu}_{z^*}.$$

The ITT regret is then

$$\mathrm{ITTRegret}(T) = \sum_{t=1}^{T} (Y_t(\chi_t(z^*)) - Y_t).$$

For every instance of the IAB problem, we can consider an underlying stochastic MAB problem given by the reward distributions $\nu_z = \sum_{\kappa \in [m]^m} p_\kappa \nu_{\kappa(z)|\kappa}$ for $z \in [m]$. The standard MAB regret in this MAB problem coincides with the ITT regret of the IAB problem. Correspondingly, we can transform any MAB algorithm to an IAB algorithm by simply recording the rewards $Y_t$ for instrument-arm pulls $Z_t$ and ignoring the in-fact applied treatment $X_t$. Then, naturally, if the original MAB algorithm has sublinear expected MAB regret guarantees then it will also have sublinear expected ITT regret guarantees because the two are identical.

When we apply this transformation to the UCB algorithm, we call the resulting algorithm UCB-ITT. An alternative transformation would have us update the estimates of the UCB algorithm as though the arm $X_t$ were pulled rather than $Z_t$, corresponding to an estimate of the mean rewards conditioned on $X_t$. We call the resulting algorithm UCB-AT for "as treated." UCB-AT has no general ITT regret guarantees but is, naturally, the same as UCB-ITT in the specific case that the IAB problem is an instance of MAB.

## 3.2 Static Treatment Regret

ITT regret completely ignores the causal effects of treatments. Minimizing ITT regret, one does not seek to select or identify the best treatment because, in particular, the benchmark is the best instrument rather than the best treatment. An alternative notion of regret we can consider in the IAB problem is the cumulative reward of a bandit algorithm relative to the cumulative reward of the best single treatment, which we call the static treatment regret. This captures the difference to the reward that would be achieved by implementing the best single treatment in the whole population as the one and *only* treatment option with no possible alternatives.

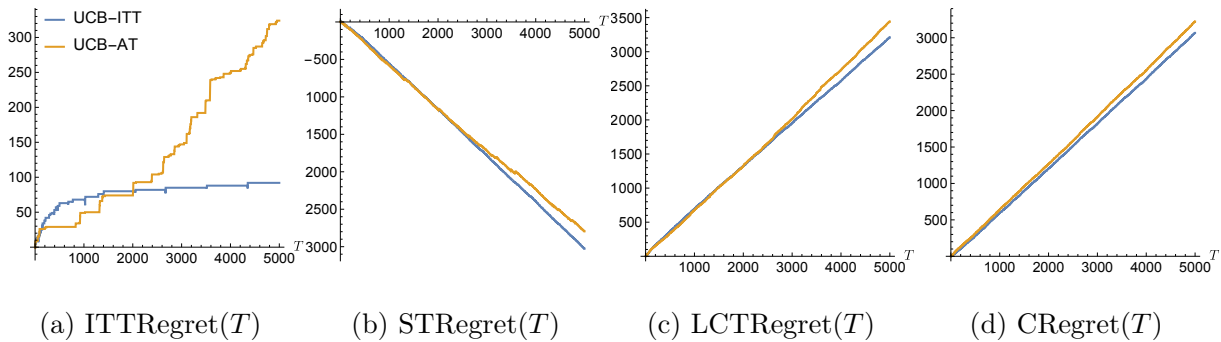

Figure 2: Various Notions of IAB Regret in Ex. 1

(a) ITTRegret(T)  (b) STRegret(T)  (c) LCTRegret(T)  (d) CRegret(T)

Let us define $x^*$ as an arbitrary choice in the set of singly-optimal treatments:

$$x^* \in \mathcal{X}^* = \text{argmax}_{x \in [m]} \mu_x, \quad \mu^* = \mu_{x^*}.$$

The static treatment regret is defined as

$$\text{STRegret}(T) = \sum_{t=1}^{T}(Y_t(x^*) - Y_t).$$

While static treatment regret corresponds to MAB regret for MAB instances, in general IAB instances it can be negative. Given a policy $\pi$, it is clear that if there is a $z \in [m]$ with $\mathbb{E}_\pi \sum_{t=1}^{T} \mathbb{I}[Z_t \neq z] = o(T)$ and $\mu^* < \tilde{\mu}_z$ then $\mathbb{E}_\pi \text{STRegret}(T) = O(-T)$. That is to say, because compliance correlates with rewards, it is, for example, feasible that even when pulling only a single instrument arm repeatedly, the compliance always forces the application of the treatment that, conditioned on the particular compliance type, does in fact have a better conditional expected reward than the the treatment that is marginally best, *i.e.*, is best on average over all types, so that $\mu^* < \tilde{\mu}_z$. An example is given below.

**Example 1.** Consider an example with $m = 3$ arms, $p_\iota = 5/8$, $p_\kappa = 3/8$ where $\kappa = (1, 1, 2)$, $Y(x) = \mu_{x|\chi} + \epsilon$, $\epsilon \sim \text{Unif}[-1, 1]$, $(\mu_{1|\iota}, \mu_{2|\iota}, \mu_{3|\iota}) = (1, -1, 0)$, and $(\mu_{1|\kappa}, \mu_{2|\kappa}, \mu_{3|\kappa}) = (-4, 0, -2)$. Then, we have that $(\mu_1, \mu_2, \mu_3) = (-7/8, -5/8, -3/4)$ so that $x^* = 2$ and $(\tilde{\mu}_1, \tilde{\mu}_2, \tilde{\mu}_3) = (-7/8, -17/8, 0)$ so that $z^* = 3$ and $\mu^* < \tilde{\mu}^*$. We consider running UCB-ITT and UCB-AT on a single sample path of the data and recording the regret accumulated. We plot the ITT regret in Fig. 2a and the static treatment regret in Fig. 2b. We note that UCB-ITT achieves logarithmic ITT regret and UCB-AT achieves linear. Both have *negative* linear static treatment regret

Because of this potential negativity, static treatment regret may be inappropriate in general for characterizing the performance of IAB algorithms and, in particular, their learning of the "best" treatment for maximal causal effect.

## 3.3 Latent-Class Treatment Regret

To derive an alternative notion of regret, we considers the IAB problem with the unobserved compliance type $\chi$ being a latent class, as in MABLC [14] but without a contextual observation. Because compliance type may correlate with rewards, we may consider our performance relative to the cumulative outcome of the best treatment for each type.

Let us define $x^*_\kappa$ as an arbitrary choice in the set of optimal treatments for unit type $\kappa$:

$$x^*_\kappa \in \mathcal{X}^*_\kappa = \mathrm{argmax}_{x \in [m]} \mu_{x|\kappa}, \quad \mu^*_{|\kappa} = \mu_{x^*_\kappa|\kappa}$$

The latent-class treatment regret is defined as

$$\mathrm{LCTRegret}(T) = \sum_{t=1}^T (Y_t(x^*_{\chi_t}) - Y_t).$$

Unlike the static treatment regret, the latent-class treatment regret is necessarily nonnegative in expectation for any bandit algorithm. On the other hand, in general settings, the latent-class treatment regret is also linear for any bandit algorithm.

**Theorem 1.** *If there exists $\ell \geq 1$ and $\kappa_1, \ldots, \kappa_\ell$ such that $p_{\kappa_1}, \ldots, p_{\kappa_\ell} > 0$, $\bigcap_{j=1}^\ell \kappa_j^{-1}(\mathcal{X}^*_{\kappa_j}) = \varnothing$ then $\mathbb{E}_\pi \mathrm{LCTRegret}(T) = \Omega(T)$ for any bandit algorithm $\pi$.*

Here $\kappa^{-1}(S) = \{z \in [m] : \kappa(z) \in S\}$ denotes the pre-image of a set $S \subset [m]$. All proofs are given in the supplemental materials.

The condition says that there exist compliance types such that no single instrument achieves optimal treatment for all types. This can happen when just a single compliance type has no instrument achieving its optimal treatment such as a 1-always-taker for which treatment 2 is optimal. More realistically, this can happen when both compliers and non-compliers exist ($p_\iota \in (0,1)$) and there is at least one non-compliance class for which no optimal instrument is also in the optimal treatment set for compliers, $\mathcal{X}^*_\iota$. So, in a general setting, treatment regret is always hopelessly linear, regardless of our algorithm. For example, in the case of Ex. 1, the latent-class treatment regret of UCB-ITT and UCB-AT is shown in Fig. 2c and is indeed linear.

Without hope of achieving sublinear regret, latent-class treatment regret is unhelpful in characterizing "good" and "bad" bandit algorithms for the IAB problem in general settings.

## 3.4 Compliers' Regret

Finally, we consider an intermediate form of regret. In the previous section we saw that because units do not always comply with treatment, we may end up with linear treatment regret when comparing to the best treatment (given compliance type). Instead, we may consider *only* the units that comply, for whom we have hope of providing any help. As we will see, under certain homogeneity assumptions, having sublinear complier's regret corresponds to bandit algorithms that learn the best treatment(s) in terms of causal effect.

The compliers' regret is defined as

$$\text{CRegret}(T) = \sum_{t=1}^{T} \mathbb{I}[\chi_t = \iota](Y_t(x_\iota^*) - Y_t)$$

There may be various situations where it is far more relevant to identify causal effects and measure regret in terms of relative treatment effect rather than focusing on ITT effect and ITT regret. For example, if the bandit problem is modeling a clinical trial, then the foremost concern is finding the best treatment and a secondary concern is good outcomes for all patients. Moreover, it may be daft to assign a particular treatment not because it has a good causal effect but because, on average, over the trial population, which includes non-compliers, it will lead to better outcomes Indeed, the population and compliance rates differ significantly between the trial stage and an implementation stage since, with enough evidence of effect, all subjects might comply. Thus, in clinical trials and other dynamic RCTs, it is more appropriate to consider treatment regret, and the compliers' regret is the appropriate treatment regret to consider as we have ruled out the alternative forms.

We begin by summarizing the (self-evident) fact that, when the IAB problem is an instance of MAB, all forms of regret we presented are equivalent:

**Theorem 2.** *If $p_\iota = 1$ then* $\text{ITTRegret}(T) = \text{STRegret}(T) = \text{LCTRegret}(T) = \text{CRegret}(T)$.

Next, we proceed to analyze compliers' regret. First, we show that, in general, one cannot have both sublinear ITT and compliers' regret.

**Assumption 1** (Compliers exist). $p_\iota = \mathbb{P}(\chi = \iota) > 0$.

**Theorem 3.** *If $\mathcal{X}_\iota^* \cap \mathcal{Z}^* = \varnothing$ and Asm. 1 holds then $\mathbb{E}_\pi \text{ITTRegret}(T) + \mathbb{E}_\pi \text{CRegret}(T) = \Omega(T)$.*

**Corollary 4.** *If $\mathbb{E}_\pi \text{ITTRegret}(T) = o(T)$, $\mathcal{X}_\iota^* \cap \mathcal{Z}^* = \varnothing$, and Asm. 1 holds then we have that $\mathbb{E}_\pi \text{CRegret}(T) = \Omega(T)$.*

The corollary shows that, in general setting, standard bandit algorithms such as UCB-ITT must achieve *linear* compliers' regret because they achieve sublinear ITT regret. That means, in particular, that they are not appropriate for learning the best treatment(s). In the case of Ex. 1, Fig. 2d shows the linear compliers' regret of UCB-ITT and UCB-AT.

However, in one simple two-armed case, we can in fact show that ITT regret is exactly identical to compliers' regret even in the presence of non-compliance:

**Theorem 5.** *If $m = 2$ and $\chi(2) \geq \chi(1)$. Then $\mathcal{X}_\iota^* = \mathcal{Z}^*$ and $\text{CRegret}(T) = \text{ITTRegret}(T)$.*

This case corresponds to the identifiable case in Angrist et al. [3]. In general settings, the identity does not hold. But in order to achieve sublinear control on compliers' regret we will need to restrict to the homogeneous setting.

## 4 Homogeneous IAB and IAB Bandit Algorithms

Next, in order to devise algorithms that can achieve sublinear compliers' regret we consider the homogeneous setting, where all *relative* treatment effects are independent of compliance type.

**Assumption 2** (Homogeneity). $Y(x) - Y(x')$ is mean-independent of $\chi$ for all $x, x' \in [m]$, *i.e.*,

$$\mathbb{E}[Y(x) - Y(x') \mid \chi] = \mathbb{E}[Y(x) - Y(x')] \quad \forall x, x' \in [m].$$

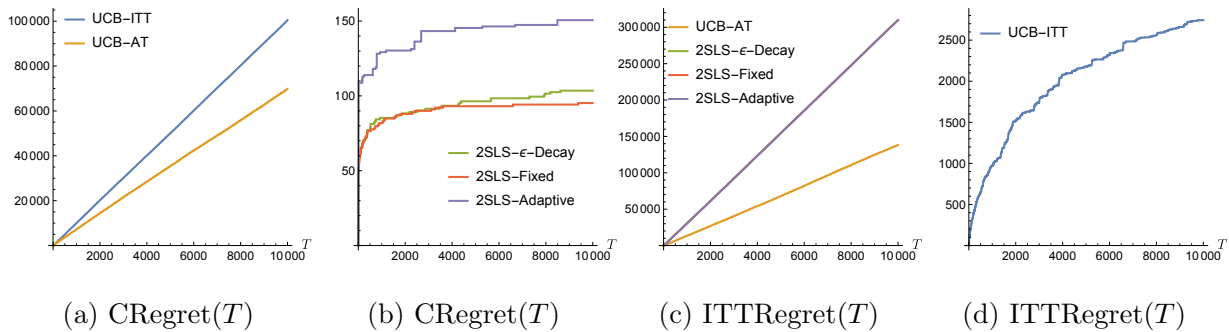

Figure 3: IAB Regret in Ex. 2

(a) CRegret(T)  (b) CRegret(T)  (c) ITTRegret(T)  (d) ITTRegret(T)

A stronger form of homogeneity, which we do not require, is that the additive model $Y(x) = \mu_x + \epsilon$ holds, which is both necessary and sufficient for $Y(x) - Y(x')$ to be constant.

Under Asm. 2, we immediately have that $\mathcal{X}^*_\kappa = \mathcal{X}^*$ for every $\kappa$. We will seek bandit algorithms that are able to identify $\mu$ and hence $\mathcal{X}^*$. Toward that end define the transition matrix

$$P_{zx} = \mathbb{P}\left(\chi(z) = x\right) = \sum_{\kappa \in [m]^m : \kappa(z) = x} p_\kappa.$$

**Theorem 6.** *Under Asm. 2, $\tilde{\mu}_z = \sum_{x=1}^m P_{zx} \mu_x$.*

Thus under homogeneity, a bandit algorithm need only estimate $\tilde{\mu}$ and $P$ in order to estimate $\mu$ via $P^{-1}\tilde{\mu}$. Next, we present two simple bandit algorithms that proceed in that way.

First, we define some statistics. We let $\mathcal{T}_z^{(t)} = \{s \leq t : Z_s = z\}$ be the set of rounds up to $t$ when $z$ was pulled, $n_z^{(t)} = |\mathcal{T}_z^{(t)}|$ the number times $z$ was pulled, $n_{zx}^{(t)} = \sum_{s \in \mathcal{T}_z^{(t)}} \mathbb{I}[X_s = x]$, the number times $z$ was pulled and $x$ was applied, $\hat{P}_{zx}^{(t)} = (n_z^{(t)})^{-1} n_{zx}^{(t)}$ the direct estimate of $P$ with data up to $t$, $\hat{\tilde{\mu}}_z^{(t)} = \frac{1}{n_z^{(t)}} \sum_{s \in \mathcal{T}_z^{(t)}} Y_s$, the direct estimate of $\tilde{\mu}$ with data up to $t$, $\hat{\mu}^{(t)} = (\hat{P}^{(t)})^{-1} \hat{\tilde{\mu}}^{(t)}$ the plug-in estimate $\mu$ arising from Thm. 6, and $\hat{z}^{(t)} \in \text{argmax}_{z=1,\ldots,m} \hat{\mu}_z^{(t-1)}$ a greedy estimate of the best treatment based on all the data available at the beginning of round $t$. The estimate $\hat{\mu}^{(t)}$ is analogous to the 2SLS procedure.

We next present an $\epsilon$-decay type of algorithm and a fixed-schedule algorithm.

---

**2SLS-$\epsilon$-decay($\alpha$)**

With probability $m/(\alpha t)$ pull a uniformly random arm and otherwise pull $\hat{z}^{(t)}$.

---

> **2SLS-fixed-schedule($\alpha$)**
> Let $z_0^{(t)} \in \text{argmin}_{z=1,\ldots,m} n_z^{(t-1)}$; if $n_{z_0^{(t)}}^{(t-1)} < \log(t)/\alpha$ then pull $z_0^{(t)}$ and otherwise pull $\hat{z}^{(t)}$.

Under certain assumptions, we can show that both of these algorithms can achieve logarithmic regret. These regret bounds are based on new concentration inequalities proven in the supplement.

**Assumption 3** (Identifiability). $P$ is invertible, i.e., $\sigma_{\min}(P) > 0$.

**Assumption 4** (Subgaussian Rewards). There exists $\psi > 0$ such that $\mathbb{E}e^{\lambda(Y(x) - \mu_x)} \leq e^{\psi \lambda^2}$ $\forall \lambda \in \mathbb{R}$.[2]

**Theorem 7.** *Suppose $2\alpha \leq \sigma_{\min}^2(P) \min\{1/(8m), \min_{x \notin \mathcal{X}^*}(\mu^* - \mu_x)^2/(16m\psi^2)\}$, $\alpha \leq 1/2$, and that Asms. 2, 3 and 4 hold. Then, for a constant $C > 0$, playing 2SLS-$\epsilon$-decay($\alpha$) has regret*

$$\mathbb{E}_\pi \text{CRegret} \leq p_\iota \sum_{x=1}^m (\mu^* - \mu_x) \log(T)/\alpha + C.$$

**Theorem 8.** *Suppose $\alpha < \sigma_{\min}^2(P) \min\{1/(8m), \min_{x \notin \mathcal{X}^*}(\mu^* - \mu_x)^2/(16m\psi^2)\}$ and that Asms. 2, 3 and 4 hold. Then, for constant $C > 0$, 2SLS-fixed-schedule($\alpha$) has regret*

$$\mathbb{E}_\pi \text{CRegret} \leq p_\iota \sum_{x=1}^m (\mu^* - \mu_x) \log(T)/\alpha + C.$$

These results, however, depend on tuning $\alpha$ based on problem-specific parameters. Nonetheless, motivated by these result, we devise a version of the fixed-schedule algorithm that adaptively estimates theses parameters and the resulting $\alpha$ but has no theoretical regret guarantees.

> **2SLS-adaptive($\gamma$)**
> Let $\hat{\delta} = \inf\{\max_{z=1,\ldots,m} \hat{\mu}_z^{(t-1)} - \hat{\mu}_{z'}^{(t-1)} > 0 : z' = 1, \ldots, m\}$ or zero if not feasible and let $\hat{\alpha}^{(t)} = \gamma \sigma_{\min}^2(\hat{P}^{(t-1)}) \hat{\delta}^2/(8m(\hat{\delta} + \sqrt{2\psi}))$. Play 2SLS-fixed-schedule($\hat{\alpha}^{(t)}$) for this round.

**Example 2.** We consider an example of a homogeneous IAB with $m = 5$. We let $Y = \mu_x + \epsilon$ where $\mu = (0.9, 1, 1, 1, 2)$ and $\epsilon \sim \mathcal{N}(0, 1)$ so that $x^* = 5$. If $\epsilon \geq -\Phi^{-1}(0.1)$ (where $\Phi$ is the normal CDF) then $\chi = \kappa^{(1)}$ and if $\epsilon \leq \Phi^{-1}(0.1)$ then $\chi = \kappa^{(5)}$. Otherwise, independently with probability $1/8$, $\chi = \kappa^{(5)}$, with probability $3/8$, $\chi = (1, 2, 3, 4, \omega)$, and with probability $1/2$, $\chi = (5, 5, 5, 5, \omega)$, where $\omega \sim \text{multinom}(1/7, 2/7, 2/7, 2/7, 0)$. This leads to $\tilde{\mu} = (1.4, 1.5, 1.5, 1.5, 1.2)$ so that $\mathcal{Z}^* = \{2, 3, 4\}$. We let $\gamma = 20$ and $\alpha^*$ be the the quantity in the bound in Thm. 8. We simulate a single sample path where we apply each of UCB-ITT, UCB-AT, 2SLS-$\epsilon$-decay($\gamma\alpha^*$), 2SLS-fixed-schedule($\gamma\alpha^*$), and 2SLS-adaptive($\gamma$). We use $\gamma = 20$ because the bounds in the theorems are conservative. We plot the resulting ITT and compliers' regrets in Fig. 3, separating between plots of algorithms that achieve linear and sublinear regret. We note that UCB-ITT achieves sublinear ITT regret but linear compliers' regret whereas our new algorithms achieve sublinear complier's regret and linear ITT regret. In particular, the heuristic adaptive policy seems to perform well without needing to know the problem-specific parameters. UCB-AT achieve linear regret in all cases.

---

[2] If $Y(x) \in [a, b]$ then Hoeffding's lemma is the statement that Asm. 4 holds with $\psi = (b-a)^2/8$. If $Y(x)$ is normal then Asm. 4 holds with $\psi$ equal to half its variance.

## 5 Conclusion

We considered a modification of the classic MAB where the arms represented mere instruments so that the assigned and applied treatment may differ. Such a setting arises when experimental units may not always comply with assigned treatments, as may be the case in any trial involving human subjects. In fact, in such trials, experience shows that noncompliance may be extremely prevalent. The possibility of noncompliance gave rise to our new IAB problem and we showed that noncompliance meant that there are a variety of possible ways to measure regret, all of which are generally different but coincide in the case of the classic MAB problem. However, by mathematically characterizing these divergent notions of regret, both for any bandit algorithm and for standard bandit algorithms that achieve sublinear ITT regret, we argued that an appropriate form of regret when one cares about the causal effect of treatments is the compliers' regret. As standard bandit algorithms achieve linear compliers' regret, we developed three new bandit algorithms that are apt for the IAB problem. We proved logarithmic regret bounds on two of these, but this depended on the algorithm being tuned to problem parameters. The third algorithm dynamically estimated these parameters but has no theoretical guarantee, although it performs well empirically.

# Proofs of IAB Regret Characterizations

*Proof of Thm. 1.* Set

$$p_{\min} = \min_{j=1,\ldots,\ell} p_{\kappa_j},$$

$$\delta = \min_{j=1,\ldots,\ell} \min_{z \notin \kappa_j^{-1}(\mathcal{X}_{\kappa_j}^*)} \mathbb{E}[Y(x_{\kappa_j}^*) - Y(\kappa_j(z)) \mid \chi = \kappa_j].$$

Then, by independence of $Z_t$, union bound, and De Morgan's law,

$$\mathbb{E}_\pi \text{LCTRegret} \geq \delta \mathbb{E}_\pi \sum_{t=1}^T \sum_{j=1}^\ell \mathbb{I}[\chi_t = \kappa_j, Z_t \notin \kappa_j^{-1}(\mathcal{X}_{\kappa_j}^*)]$$

$$= \delta \sum_{t=1}^T \sum_{j=1}^\ell p_{k_j} \mathbb{P}_\pi(Z_t \notin \kappa_j^{-1}(\mathcal{X}_{\kappa_j}^*))$$

$$\geq \delta p_{\min} \sum_{t=1}^T \sum_{j=1}^\ell \mathbb{P}_\pi(Z_t \notin \kappa_j^{-1}(\mathcal{X}_{\kappa_j}^*))$$

$$\geq \delta p_{\min} \sum_{t=1}^T \mathbb{P}_\pi(\vee_{j=1}^\ell Z_t \notin \kappa_j^{-1}(\mathcal{X}_{\kappa_j}^*))$$

$$= \delta p_{\min} \sum_{t=1}^T (1 - \mathbb{P}_\pi(Z_t \in \varnothing))$$

$$= \delta p_{\min} T = \Omega(T) \qquad \square$$

*Proof of Thm. 3.* Define $\delta' = \min_{z \in \mathcal{X}_\iota^*} \mathbb{E}[Y(\chi(z^*)) - Y(\chi(z))]$, $\delta'' = \min_{z \notin \mathcal{X}_\iota^*} \mathbb{E}[Y(x_\iota^*) - Y(z) \mid \chi = \iota]$, $\delta = \min\{\delta', \delta''\}$, and $T_{\mathcal{X}_\iota^*} = \sum_{t=1}^T \mathbb{I}[Z_t \in \mathcal{X}_\iota^*]$. By assumption, we have $\delta > 0$. By independence of $Z_t$, we have

$$\mathbb{E}_\pi \text{ITTRegret} \geq \mathbb{E} \sum_{t=1}^T \mathbb{I}[Z_t \in \mathcal{X}_\iota^*](Y_t(\chi_t(z^*)) - Y_t)$$

$$\geq \delta' \mathbb{E} T_{\mathcal{X}_\iota^*} \geq \delta \mathbb{E} T_{\mathcal{X}_\iota^*}.$$

By independence of $Z_t$, we also have

$$\mathbb{E}_\pi \text{CRegret} \geq \mathbb{E} \sum_{t=1}^T \mathbb{I}[\chi_t = \iota, Z_t \notin \mathcal{X}_\iota^*](Y_t(x_\iota^*) - Y_t)$$

$$\geq \delta'' \mathbb{E} \sum_{t=1}^T \mathbb{I}[\chi_t = \iota, Z_t \notin \mathcal{X}_\iota^*]$$

$$\geq \delta \mathbb{E} \sum_{t=1}^{T} (\mathbb{I}[\chi_t = \iota] - \mathbb{I}[Z_t \in \mathcal{X}_\iota^*])$$

$$= \delta(p_\iota T - \mathbb{E} T_{\mathcal{X}_\iota^*}).$$

Together, $\mathbb{E}_\pi \text{ITTRegret} + \mathbb{E}_\pi \text{CRegret} = \Omega(T)$. □

*Proof.* By assumption $\chi(2) \geq \chi(1)$, we must have $\chi \in \{\iota, \kappa^{(1)}, \kappa^{(2)}\}$. Therefore,

$$\mathbb{E}[Y(\chi(z))] = \mathbb{E}[\mathbb{E}[Y(\chi(z)) \mid \chi]]$$
$$= p_\iota \mathbb{E}[Y(z) \mid \chi = \iota] + p_{\kappa^{(1)}} \mathbb{E}[Y(1) \mid \chi = \kappa^{(1)}] + p_{\kappa^{(2)}} \mathbb{E}[Y(2) \mid \chi = \kappa^{(2)}].$$

Maximizing both sides over $z \in [m]$, we see that $\mathcal{Z}^* = \mathcal{X}_\iota^*$. Therefore,

$$\text{ITTRegret} = \sum_{t=1}^{T} \mathbb{I}[\chi_t = \iota](Y_t(\chi_t(z^*)) - Y_t)$$
$$+ \sum_{t=1}^{T} \mathbb{I}[\chi_t = \kappa^{(1)}](Y_t(\chi_t(z^*)) - Y_t)$$
$$+ \sum_{t=1}^{T} \mathbb{I}[\chi_t = \kappa^{(2)}](Y_t(\chi_t(z^*)) - Y_t)$$
$$= \text{CRegret}$$
$$+ \sum_{t=1}^{T} \mathbb{I}[\chi_t = \kappa^{(1)}](Y_t(1) - Y_t(1))$$
$$+ \sum_{t=1}^{T} \mathbb{I}[\chi_t = \kappa^{(2)}](Y_t(2) - Y_t(2))$$
$$= \text{CRegret} \qquad □$$

*Proof of Thm. 6.* Fix $x' \in [m]$. By Asm. 2 we have $\mathbb{E}[Y(x) - Y(x') \mid \chi(z) = x] = \mu_x - \mu_{x'}$. Therefore:

$$\tilde{\mu}_z = \mathbb{E}[Y(\chi(z))] = \sum_{x=1}^{m} P_{zx} \mathbb{E}[Y(x) \mid \chi(z) = x]$$
$$= \sum_{x=1}^{m} P_{zx}(\mu_x - \mu_{x'} + \mathbb{E}[Y(x') \mid \chi(z) = x])$$
$$= \sum_{x=1}^{m} P_{zx}\mu_x - \mu_{x'} + \mu_{x'} = \sum_{x=1}^{m} P_{zx}\mu_x \qquad □$$

# Proofs of Regret Bounds

**Lemma 1.** $\mathbb{P}(\|\hat{P}^{(t)} - P\|_\infty > \epsilon \mid Z_1, \ldots, Z_t) \leq \sum_{z=1}^{m} e^{m - \frac{1}{8} n_z^{(t)} \epsilon^2}$

*Proof.* We condition on $Z_1, \ldots, Z_t$ and treat them as fixed *throughout* the argument. Define $e_z \in \mathbb{R}^m$ as the one-hot unit vector with one in coordinate $z$. Let us define the empirical Rademacher complexities

$$\widehat{\mathfrak{R}}_z^{(t)} = \frac{1}{2^{n_z^{(t)}}} \sum_{\xi \in \{-1,+1\}^{\mathcal{T}_z^{(t)}}} \sup_{\|v\|_\infty \leq 1} \frac{1}{n_z^{(t)}} \sum_{s \in \mathcal{T}_z^{(t)}} \xi_s v^T e_{X_s}.$$

Note that

$$\|\hat{P}_z^{(t)} - P_z\|_1 = \sup_{\|v\|_\infty \leq 1} \left( \frac{1}{n_z^{(t)}} \sum_{s \in \mathcal{T}_z^{(t)}} v^T e_{X_s} - v^T P_z \right)$$

Then by Bartlett and Mendelson [6], we have that with probability at least $1 - \eta$,

$$\|\hat{P}_z^{(t)} - P_z\|_1 \leq 2 \mathbb{E}[\widehat{\mathfrak{R}}_z^{(t)}] + \sqrt{\log(1/\eta)/(2 n_z^{(t)})}$$

By linearity and duality of norms, we have

$$\widehat{\mathfrak{R}}_z^{(t)} = \frac{1}{2^{n_z^{(t)}}} \sum_{\xi \in \{-1,+1\}^{n_z^{(t)}}} \left\| \frac{1}{n_z^{(t)}} \sum_{s=1}^{n_z^{(t)}} \xi_s e_{X_s} \right\|_1$$

$$= \frac{1}{n_z^{(t)}} \sum_{x=1}^{m} \frac{1}{2^{n_z^{(t)}}} \sum_{\xi \in \{-1,+1\}^{n_z^{(t)}}} \left| \sum_{s=1}^{n_z^{(t)}} \mathbb{I}[X_s = x] \xi_t \right|$$

$$= \frac{1}{n_z^{(t)}} \sum_{x=1}^{m} \frac{1}{2^{n_{zx}^{(t)}}} \sum_{\xi \in \{-1,+1\}^{n_{zx}^{(t)}}} \left| \sum_{s=1}^{n_{zx}^{(t)}} \xi_s \right|$$

$$= \frac{1}{n_z^{(t)}} \sum_{x=1}^{m} \frac{1}{2^{n_{zx}^{(t)}-1}} \left\lceil \frac{n_{zx}^{(t)}}{2} \right\rceil \binom{m}{\left\lceil \frac{n_{zx}^{(t)}}{2} \right\rceil}$$

$$\leq \frac{1}{n_z^{(t)}} \sum_{x=1}^{m} \sqrt{n_{zx}^{(t)}} = \frac{1}{\sqrt{n_z^{(t)}}} \sum_{x=1}^{m} \sqrt{\hat{P}_{zx}^{(t)}} \leq \sqrt{\frac{m}{n_z^{(t)}}}.$$

Therefore, by $2 \geq 1/\sqrt{2}$, Jensen's inequality, and the concavity of square root, with probability at least $1 - \eta$,

$$\|\hat{P}_z^{(t)} - P_z\|_1 \leq 4 \sqrt{(m - \log \eta)/(2 n_z^{(t)})}.$$

Finally, $\mathbb{P}(\|\hat{P}^{(t)} - P\|_\infty > \epsilon) = \mathbb{P}(\exists z : \|\hat{P}_z^{(t)} - P_z\|_1 > \epsilon)$ and the union bound complete the proof. $\square$

The Rademacher complexity argument (compared to an argument based on Hoeffding inequality's and the union bound) is critical to achieving the correct rate in $n_z^{(t)}/m$.

**Lemma 2.** *Let $\xi \in (0,1)$ and $0 < \underline{\sigma} \leq \sigma_{\min}(P)$.*

$$\mathbb{P}(\|(\hat{P}^{(t)})^{-1}\|_\infty > \sqrt{m}\underline{\sigma}^{-1}(1-\xi)^{-1} \mid Z_1, \ldots, Z_t) \leq \sum_{z=1}^m e^{m - \frac{1}{8}n_z^{(t)}\underline{\sigma}^2\xi^2/m}$$

*Proof.* We have

$$\|(\hat{P}^{(t)})^{-1}\|_\infty \leq \sqrt{m}\|(\hat{P}^{(t)})^{-1}\|_2 = \frac{\sqrt{m}}{\sigma_{\min}(\hat{P}^{(t)})}$$
$$\leq \frac{\sqrt{m}}{(\underline{\sigma} - \|\hat{P}^{(t)} - P\|_2)_+} \leq \frac{1}{(\underline{\sigma}/\sqrt{m} - \|\hat{P}^{(t)} - P\|_\infty)_+}.$$

Applying Lemma 1 with $\epsilon = \underline{\sigma}\xi/\sqrt{m}$ yields the result. $\square$

**Lemma 3.** *Let Asm. 4 hold. Then,*

$$\mathbb{P}(|\hat{\tilde{\mu}}_z^{(t)} - \hat{P}_z^{(t)}\mu| > \epsilon \mid Z_1, \ldots, Z_t) \leq 2e^{-n_z^{(t)}\epsilon^2/(4\psi)}.$$

*Proof.* Note that we can rewrite

$$\hat{\tilde{\mu}}_z^{(t)} - \hat{P}_z^{(t)}\mu = \tfrac{1}{n_z^{(t)}} \sum_{s \in \mathcal{T}_z^{(t)}} (Y_t(X_t) - \mu_{X_t}).$$

By Markov's inequality,

$$\mathbb{P}(\hat{\tilde{\mu}}_z^{(t)} - \hat{P}_z^{(t)}\mu > \epsilon \mid Z_1, X_1, \ldots, Z_t, X_t)$$
$$\leq \inf_{\lambda \geq 0} e^{-n_z^{(t)}\lambda\epsilon} \prod_{s=1}^t \mathbb{E}[e^{\lambda(Y_s(X_s) - \mu_{X_s})} \mid X_s]$$
$$\leq \inf_{\lambda \geq 0} e^{n_z^{(t)}(\psi\lambda^2 - \lambda\epsilon)} = e^{-n_z^{(t)}\epsilon^2/(4\psi)}.$$

A symmetric argument, the union bound, and marginalizing over $X_1, \ldots, X_t$ yields the result. $\square$

**Lemma 4.** *Let $0 < \underline{\sigma} \leq \sigma_{\min}(P)$ and Asm. 4 hold. Then,*

$$\mathbb{P}(\|\hat{\mu}^{(t)} - \mu\|_\infty > \epsilon \mid Z_1, \ldots, Z_t) \leq 2\sum_{z=1}^m e^{m - \frac{n_z^{(t)}\underline{\sigma}^2\epsilon^2}{2m(2\epsilon + \sqrt{2\psi})^2}}$$

*Proof.* Applying Lemma 2 with $\xi = 2\epsilon/(2\epsilon + \sqrt{2\psi})$, applying Lemma 3 with $\epsilon(1-f)\underline{\sigma}^{-1}m^{-1/2}$ for each $z$, the union bound twice, and $2 \leq e^m$ yield the result. $\square$

*Proof of Thm. 7.* Let $E^{(t)} = \sum_{s=1}^{t} \epsilon_s$, $B = \max_{x=1,\ldots,m}(\mu^* - \mu_x)$, and $\delta = \min_{x \notin \mathcal{X}^*}(\mu^* - \mu_x)$. Then,

$$\mathbb{E}_\pi \text{CRegret}(T) \leq p_\iota \sum_{x=1}^{m}(\mu^* - \mu_x)E^{(T)}/m + B \sum_{t=1}^{\infty} \mathbb{P}(\|\hat{\mu}^{(t-1)} - \mu\|_\infty \geq \delta/2).$$

Note that $E^{(t-1)} \geq m\log(t)/\alpha$ and $E^{(t)} \leq m(\log(t)+1)/\alpha$ so that the first regret term is $\leq \sum_{x=1}^{m}(\mu^* - \mu_x)(\log(T)+1)/\alpha$. Let $\phi = \sigma_{\min}^2(P)\delta^2/(8m(\delta - \sqrt{2\psi})^2) > 0$ and $\Phi = 1 - e^{-\phi} > 0$. By Lemma 4 and the law of total probability, $\mathbb{P}(\|\hat{\mu}^{(t)} - \mu\|_\infty \geq \delta/2) \leq 2e^m \sum_{z=1}^{m} \mathbb{E}[e^{-\phi n_z^{(t)}}]$. Let $\Gamma_t$ be one if at time step $t$ we explored (random pull) and zero otherwise. Let $\tilde{n}_z^{(t)} = \sum_{s \in \mathcal{T}_z^{(t)}} \Gamma_s \leq n_z^{(t)}$. Then

$$\log \mathbb{E}_\pi[e^{-\phi n_z^{(t)}}] \leq \log \mathbb{E}_\pi[e^{-\phi \tilde{n}_z^{(t)}}]$$
$$= \sum_{s=1}^{t} \log(e^{-\phi}\epsilon_t/m + (1 - \epsilon_t/m))$$
$$\leq -\sum_{s=1}^{t} \Phi\epsilon_t/m = -\Phi E^{(t)}/m.$$

Finally, $\sum_{t=1}^{\infty} e^{-\Phi E^{(t-1)}/m} \leq \sum_{t=1}^{\infty} t^{-\Phi/\alpha}$ and by assumption $0 < \alpha \leq \min(\phi, 1)/2 < \Phi$ so that the second regret term is a finite constant. $\square$

*Proof of Thm. 8.* Let $B = \max_{x=1,\ldots,m}(\mu^* - \mu_x)$, $\delta = \min_{x \notin \mathcal{X}^*}(\mu^* - \mu_x)$, $\Gamma_{tz} = \mathbb{I}[n_{z_0^{(t)}}^{(t-1)} < \log(t)/\alpha, z = z_0^{(t)}]$, $\tilde{n}_z^{(t)} = \sum_{s=1}^{t} \Gamma_{tz}$, and $\Gamma_t = \sum_{z=1}^{m} \Gamma_{tz}$. We then have

$$\mathbb{E}_\pi \text{CRegret}(T) \leq p_\iota \sum_{x=1}^{m}(\mu^* - \mu_x)\mathbb{E}_\pi \tilde{n}_x^{(T)} + B \sum_{t=1}^{\infty} \mathbb{P}(\|\hat{\mu}^{(t-1)} - \mu\|_\infty \geq \delta/2, \Gamma_t = 0).$$

For the first regret term, we have $\tilde{n}_x^{(T)} \leq 1 + \log(T)/\alpha$ under $\pi$ by construction and so the same is true for $\mathbb{E}_\pi \tilde{n}_x^{(T)}$. For the second term, we invoke Lemma 4 and note that whenever $\Gamma_t = 0$ we must also have $\min_{z=1,\ldots,m} n_z^{(t-1)} \geq \log t/\alpha$ in order to get

$$\mathbb{P}(\|\hat{\mu}^{(t-1)} - \mu\|_\infty \geq \delta/2, \Gamma_t = 0) \leq 2e^m m t^{-\frac{\sigma_{\min}^2(P)\delta^2}{8\alpha m(\delta + \sqrt{2\psi})^2}}.$$

By assumption, we have that $\frac{\sigma_{\min}^2(P)\delta^2}{8\alpha m(\delta + \sqrt{2\psi})^2} > 1$ so that the second regret term is a finite constant. $\square$